\documentclass[letterpaper, 10 pt, conference]{ieeeconf}  

\IEEEoverridecommandlockouts                              

\usepackage{multicol}
\usepackage[bookmarks=true]{hyperref}
\usepackage{amsfonts}
\usepackage{amsmath}
\usepackage{graphicx}
\usepackage{float}
\usepackage{caption}
\usepackage{subfigure}
\usepackage{graphics} 
\usepackage{epsfig} 
\usepackage{times} 
\usepackage{amssymb}  
\usepackage{multirow}
\usepackage{multicol}
\usepackage{hyperref}
\usepackage{tabularx}
\usepackage{hyperref}
\usepackage[usenames,dvipsnames,table]{xcolor}
\hypersetup{
    colorlinks=true,
    linkcolor=MidnightBlue,
    filecolor=magenta,      
    urlcolor=MidnightBlue,
    citecolor=MidnightBlue,
} 
\usepackage{color}
\usepackage{colortbl}

\usepackage[labelfont=bf]{caption}

\definecolor{Gray}{rgb}{0.93, 0.93, 0.93}
\newcolumntype{a}{>{\columncolor{Gray}}c}

\usepackage{booktabs}
\usepackage{etoolbox}
\usepackage{biblatex}
\makeatletter
\patchcmd{\@makecaption}
  {\scshape}
  {}
  {}
  {}
\makeatletter
\patchcmd{\@makecaption}
  {\\}
  {:\ }
  {}
  {}
\makeatother

\title{\LARGE \bf
Dexterous Manipulation Based on Prior Dexterous Grasp Pose Knowledge}

\author{Hengxu Yan, Haoshu Fang, Cewu Lu
\thanks{{Hengxu Yan, Haoshu Fang, Cewu Lu are with Shanghai Jiao Tong University, email:
        {\tt\small HengxuYan@sjtu.edu.cn, fhaoshu@gmail.com, lucewu@sjtu.edu.cn }}%
}}

\begin{document}

\maketitle
\thispagestyle{empty}
\pagestyle{empty}

\begin{abstract}
Dexterous manipulation has received considerable attention in recent research. Predominantly, existing studies have concentrated on reinforcement learning methods to address the substantial degrees of freedom in hand movements. Nonetheless, these methods typically suffer from low efficiency and accuracy. In this work, we introduce a novel reinforcement learning approach that leverages prior dexterous grasp pose knowledge to enhance both efficiency and accuracy. Unlike previous work, they always make the robotic hand go with a fixed dexterous grasp pose, We decouple the manipulation process into two distinct phases: initially, we generate a dexterous grasp pose targeting the functional part of the object; after that, we employ reinforcement learning to comprehensively explore the environment. Our findings suggest that the majority of learning time is expended in  identifying the appropriate initial position and selecting the optimal manipulation viewpoint. Experimental results demonstrate significant improvements in learning efficiency and success rates across four distinct tasks.

\end{abstract}

\section{INTRODUCTION}
Human interactions with the physical world are heavily reliant on hand movements. Observations of infant learning reveal that they initially form a preliminary manipulation position and viewpoint after brief observation. They then actively interact with an object's functional parts, continuously collecting visual feedback, which accelerates their grasp of effective manipulation strategies.
However, replicating this human-like manipulation in robots remains a significant challenge due to the high degrees of freedom in robotic hands. Recent studies ~\cite{bao2023dexart, qin2023dexpoint} have made encouraging progress in addressing this issue, yet they typically initiate the manipulation process from a fixed, human-designed position tailored to specific tasks. This approach often leads to inefficiencies, as it wastes valuable learning time exploring an extensive environmental space to determine the optimal position and viewpoint, especially when the object is randomly placed within the workspace.
~\cite{qin2022one} introduces an imitation learning method that replicates expert policies. However, this approach requires extensive data, typically collected through human teleoperation of the robotic hand, which is both time-consuming and costly. To address the challenges of data collection,  ~\cite{wang2024dexcap, arunachalam2023holo} propose dexterous manipulation data collection systems. Building on this, ~\cite{haldar2023teach} develops a rapid imitation learning approach capable of acquiring robust visual skills from less than a minute of human demonstrations. This method updates policies based on a weak initial policy that can only adapt to a limited range of robotic hand motions.

\begin{figure}[!t]
\centering
\includegraphics[width=0.45\textwidth]{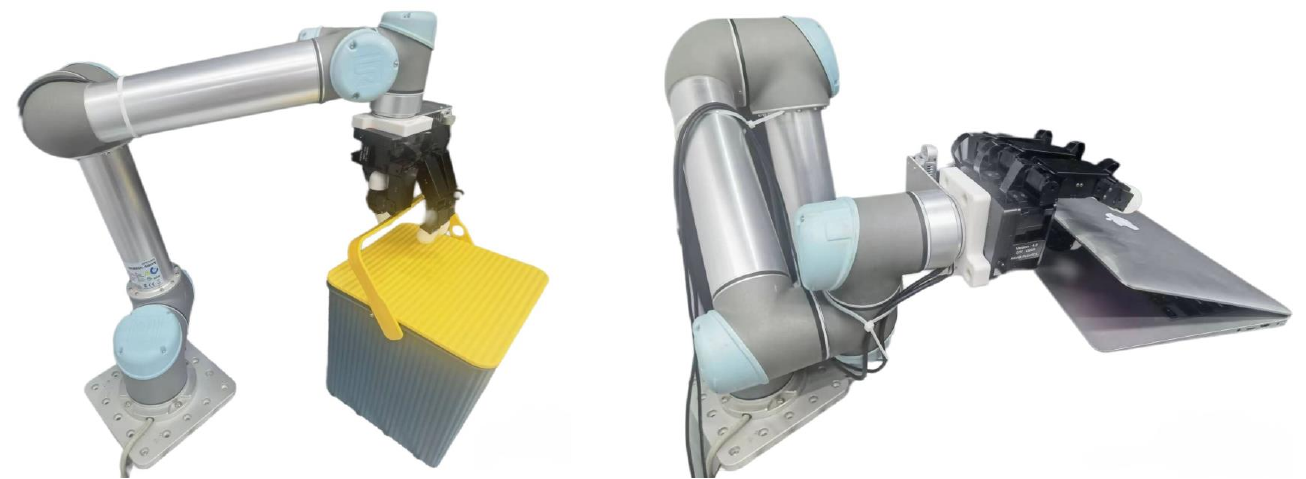}\\
\caption{For the tasks of lifting the bucket and opening the laptop, we set the initial dexterous grasp pose to facilitate successful task completion.}
\label{first_image}
\end{figure}
Inspired by the learning process observed in infants, we propose a dexterous manipulation method grounded in prior dexterous grasp pose knowledge, as illustrated in  Fig.\ref{first_image}. Our approach is structured into two phases. First, we segment the functional part of the object, which is then used to generate a set of two-finger grasp poses using Anygrasp~\cite{fang2023anygrasp}. These two-finger grasp poses are mapped into a dexterous grasp space, and a final dexterous grasp pose is selected through collision detection, serving as the initial grasp pose. This process mirrors how infants determine the optimal viewpoint and position for manipulating an object. In the second phase, we employ the Proximal Policy Optimization (PPO) algorithm ~\cite{schulman2017proximal} to iteratively interact with the object using partial-view point cloud inputs.

The principal contributions of our study are as follows:
\begin{itemize}
    \item We introduce a two-stage framework that decouples dexterous manipulation into two phases: generating an initial dexterous grasp pose for the functional part of the object and training a policy to refine this dexterous grasp pose for effective object manipulation through reinforcement learning.

    \item We decompose the dexterous manipulation reward in reinforcement learning into three distinct components: interaction reward, completion reward and restriction reward, integrating them into the learning process to facilitate task completion.
    
    \item We conduct extensive experiments to validate that leveraging prior dexterous grasp pose knowledge significantly enhances the learning efficiency and success rate of reinforcement learning in simulation environments. Furthermore, we demonstrate the successful transfer of our algorithm to real-world applications, confirming its practical performance.
\end{itemize}

\section{RELATED WORK}
 
\subsection{Dexterous manipulation}

Human dexterity is centered on the hand, and the advancement of human civilization is deeply intertwined with hands-on exploration of the surrounding environment. In this context, several studies \cite{ gupta2021reset, garcia2020physics} have employed reinforcement learning to teach robots how to manipulate objects within large spaces. However, these approaches often require training over millions of time steps to achieve optimal performance and typically rely on a fixed grasp pose. Other research \cite{chen2023visual, khandate2023sampling, allshire2022transferring, andrychowicz2020learning, nagabandi2020deep, yin2023rotating} has focused on in-hand manipulation, where the object is reoriented to a default pose using visual or tactile information. While these methods demonstrate finger dexterity, they often limit the manipulation to objects held within the palm, restricting the range of potential applications.
Several studies \cite{shao2021concept2robot, schmeckpeper2020reinforcement, ebert2021bridge, smith2019avid, chen2021learning, alakuijala2023learning} have employed human video demonstrations to teach robots object manipulation. However, significant differences between robotic and human hands, along with the diversity of robotic hand designs, pose challenges for direct transfer of these techniques. In response, some works \cite{johns2021coarse, zhang2018deep, qin2022one, wang2024dexcap} have utilized teleoperation to collect data, generating expert-level datasets by having robots manipulate real objects. Additionally, \cite{rajeswaran2017learning, arunachalam2023dexterous} propose a virtual reality (VR) setup for collecting dexterous manipulation demonstrations, introducing the Demo Augmented Policy Gradient algorithm for imitation learning. Despite these advancements, such methods often face inefficiencies in data collection and exhibit discrepancies between the collected data and real-world applications. Moreover, imitation learning is inherently limited by the capabilities of the demonstrator, as it lacks an explicit understanding of task success.

To address these shortcomings, some studies  \cite{rajeswaran2017learning, jeong2020learning, christen2019guided} have combined reinforcement learning with imitation learning to enhance learning and exploration efficiency. \cite{wu2023learning} further leverages human grasp affordances to teach dexterous manipulation. However, most existing methods either focus on learning a policy tailored to a single object or rely on access to object states provided by a perfect detector, which presents additional challenges for Sim2Real transfer.
\subsection{Dexterous Grasping}

Grasping tasks are foundational to dexterous manipulation, as they provide crucial prior knowledge about the objects being manipulated by a robotic hand. Previous research \cite{fang2023anygrasp, fang2020graspnet, Wang_2021_ICCV, ten2017grasp, mahler2019learning} has predominantly focused on two-finger grasping tasks. However, with the increasing demand for precision manipulation, other studies \cite{mandikal2022dexvip, mandikal2021learning, kokic2020learning, shaw2023videodex, 8967960} have utilized human grasp demonstrations or object affordances to teach robots how to grasp objects using dexterous hands.\cite{xu2023unidexgrasp, wan2023unidexgrasp++} employ reinforcement learning and distillation methods to develop dexterous grasping policies. However, there has yet to be an application that effectively leverages prior dexterous grasp pose knowledge for object manipulation. In this paper, we propose a dexterous manipulation method that utilizes prior dexterous grasp pose knowledge to determine the initial approach direction and grasp position, thereby enabling the robot to complete tasks more quickly and accurately.

\section{METHOD}
In this section, we address the challenge of improving learning efficiency while maintaining human-like object manipulation. Controlling each joint of a dexterous hand is inherently difficult due to the hand’s high degrees of freedom, which significantly expands the exploration space and diminishes learning efficiency. Additionally, deploying the learned policy in real-world scenarios is crucial, particularly in ensuring the safety of the manipulation process. Improper force application during object manipulation may trigger emergency stops, further complicating real-world deployment.

\begin{figure*}[!t]
\centering
\includegraphics[width=0.98\textwidth]{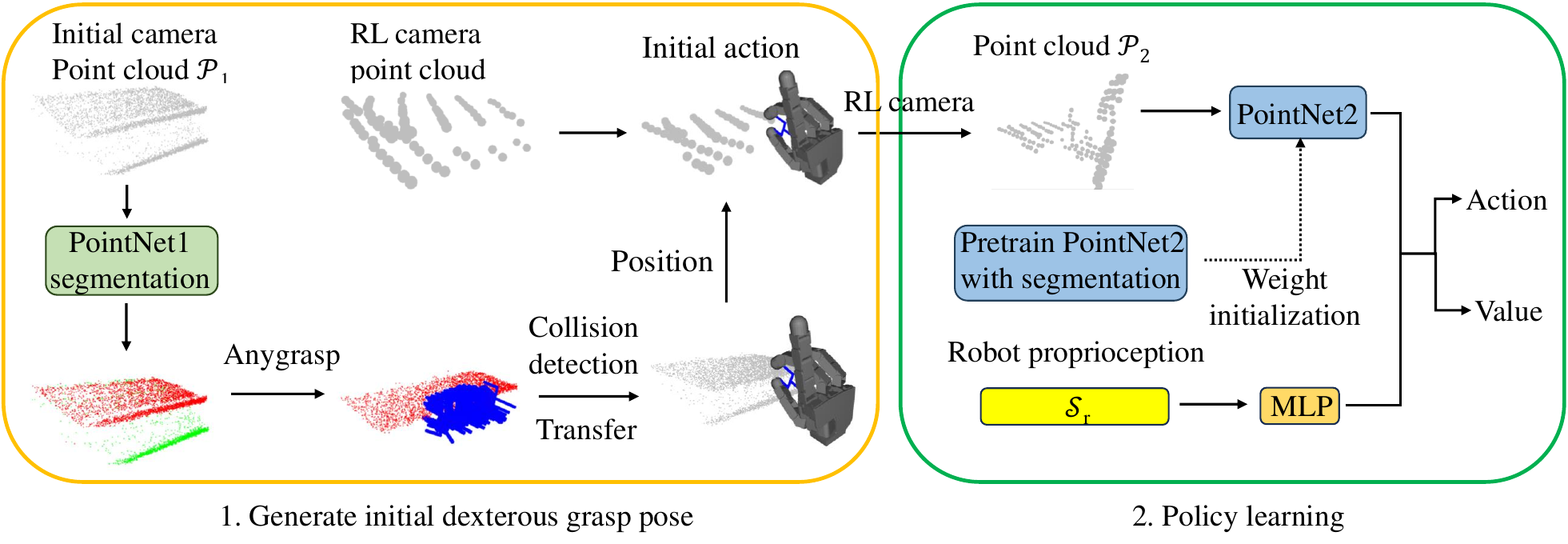}\\
\caption{Illustration of our dexterous manipulation method. We employ PPO to teach the dexterous hand how to manipulate objects based on a dexterous grasp pose. (1) Starting with a partial-view point cloud captured by the initial camera, we use PointNet1 to segment the functional part of the object, which is then used to generate a set of two-finger grasp poses with Anygrasp. These poses are subsequently mapped to a dexterous grasp space, where collision detection is applied to select an appropriate dexterous grasp pose. (2) PointNet2 serves as our backbone to extract features from the partial-view point cloud obtained by the RL camera. The backbone is pre-trained on a segmentation network before being used in RL training.}
\label{pipeline}
\end{figure*}

\subsection{Problem Formulation}

In this study, we adopt the four tasks outlined by DexArt~\cite{bao2023dexart}, which provide a benchmark for generalizable dexterous manipulation with articulated objects. An overview of our method is illustrated in Fig.~\ref{pipeline}. The first step involves generating the initial dexterous grasp pose using the point cloud $\mathcal{P}_1$ obtained from the initial camera. This point cloud $\mathcal{P}_1$ is fed into a segmentation network $\mathcal{S}_e$  to extract the point cloud of the functional part $\mathcal{P}_f$ as represented by:

\begin{equation}\label{seg formulation}
\mathcal{P}_f = \mathcal{S}_e(\mathcal{P}_1).
\end{equation}


Next, $\mathcal{P}_f$ is processed by Anygrasp ~\cite{fang2023anygrasp} to generate a set of two-finger grasp poses $\{\hat{\mathcal{G}_1},\hat{\mathcal{G}_2},\cdots, \hat{\mathcal{G}_n}\}$, denoted as:

\begin{equation}\label{anygrasp formulation}
\{\hat{\mathcal{G}_1},\hat{\mathcal{G}_2},\cdots, \hat{\mathcal{G}_n}\} = Anygrasp(\mathcal{P}_f).
\end{equation}


Subsequently, each two-finger grasp pose $\hat{\mathcal{G}_i}$ is mapped to a dexterous grasp pose $\mathcal{G}_i$ as shown in Fig.~\ref{coordinate}. A collision detector is then used to filter out any invalid dexterous grasp poses from $\mathcal{P}_1$. Finally, the pose closest to the camera is selected as the initial dexterous grasp pose $\mathcal{G}$. This approach bypasses the reinforcement learning process typically used to determine the viewpoint and position for manipulation, significantly reducing the complexity of the model’s learning tasks.


The second step involves the exploration process of the dexterous hand using reinforcement learning from the initial grasp pose $\mathcal{G}$. We first utilize IKFast~\cite{diankov_thesis} to compute the initial angle of each robotic joint. Following this, we apply PPO to adjust $\mathcal{G}$ in order to complete the task. We model this process as a Markov Decision Process (MDP) as defined by~\cite{cassandra1998survey}, represented as a 6-tuple $\left<S, A, O, R^\prime, T, U\right>$, where $S$ and $A$ denote the state and action space, $O$ is the observation space, $R^\prime(s,a)$ is the reward function, $T(s_{t+1}|s_t,a_t)$ is the transition dynamics of state $s_{t+1}$ at step $t+1$ given action $a_t$ made under state $s_t$, and $U$ is the measurement function generating observations from the state. Our goal is to maximize the expected reward with the policy $\pi(a|s)$.

\subsection{Point Cloud Segmentation}



Our segmentation dataset is generated using the SAPIEN physical simulator ~\cite{xiang2020sapien}. Objects are randomly placed within the simulation environment, and their handle positions correspond to varying degrees of task completion. Point cloud observations are then rendered from the perspective of the initial camera, simulating manipulation tasks. Fig.~\ref{target_and_segmentation_point_cloud} illustrates the point clouds for four different tasks. Our dataset comprises 3,000 point clouds for each object, with each point cloud containing both functional and non-functional parts. We then use PointNet~\cite{qi2017pointnet} to train our segmentation network, employing the GELU activation function and optimizing with CrossEntropy loss.

\subsection{Transfer the $\hat{\mathcal{G}}$ to $\mathcal{G}$}

We firstly define a dexterous grasp pose $\mathcal{G}$ as 

\begin{equation}\label{multi-finger definition1}
\mathcal{G} = [\mathbf{R}\ \mathbf{t}\ \mathbf{B}],
\end{equation}
where $\mathbf{R} \in \mathbb{R}^{3\times1} $ and
$\mathbf{t} \in \mathbb{R}^{3\times1}$ represent the rotation and translation of the end link of a robotic arm, $\mathbf{B} \in \mathbb{R}^{16\times1}$ denotes the degrees of freedom (DoF) of the dexterous hand (Allegro Hand). 

Similarly, we define a two-finger grasp pose $\hat{\mathcal{G}}$ as 
\begin{equation}\label{two-finger definition1}
\hat{\mathcal{G}} = [\mathbf{R}\ \mathbf{t}\ w],
\end{equation}
where $\mathbf{R}$ and $\mathbf{t}$ have the same meaning as in the definition of Eqn.~\ref{multi-finger definition1}, the $w$ is minimum grasp width of the two-finger. The key difference between the two definitions lies in the degrees of freedom. To map from the $w$ to the $\mathbf{B}$, we construct a mapping function $f(\cdot)$. We first discretize the continuous variable $w$ into $n$ widths and manually design four $\{\mathbf{B}_{i1}, \mathbf{B}_{i2}, \mathbf{B}_{i3}, \mathbf{B}_{i4}\}$ for every $w_i, i \in \{1,2,\cdots, n\}$  and denoted as

\begin{figure}[!t]
\centering
\includegraphics[width=0.45\textwidth]{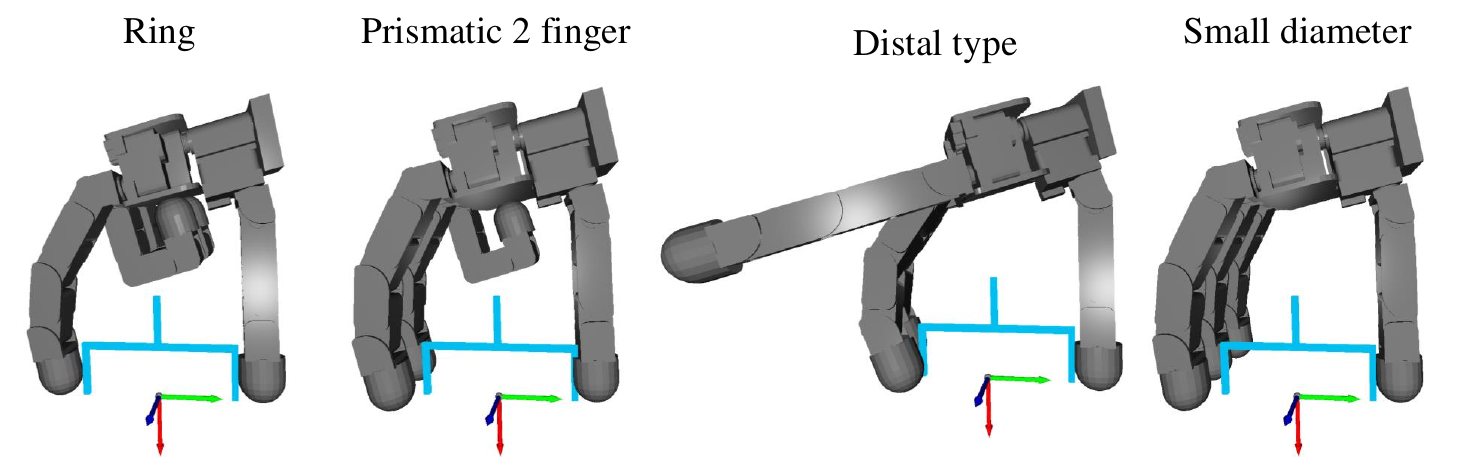}\\
\caption{Mapping of the coordinate system from two-finger grasp poses to four grasp types for the dexterous hand.}
\label{coordinate}
\end{figure}

\begin{equation}\label{two-finger definition2}
\{\mathbf{B}_{i1}, \mathbf{B}_{i2}, \mathbf{B}_{i3}, \mathbf{B}_{i4}\} = f(w_i).~~~ i \in \{1,2,\cdots, n\}.
\end{equation}
Fig.~\ref{coordinate} illustrates a mapping from the $w_i$ to each $\mathbf{B}_{ij}$.

\subsection{Reinforcement Learning}

Once the initial dexterous grasp pose is established, the viewpoint and position of the robotic arm's approach to the object's manipulated part are determined, along with the angle of each joint in the dexterous hand. Following this, we utilize PPO to refine the initial dexterous grasp pose. In our algorithm, we extract features from both the point cloud and the robot's proprioception. These features are then used to generate the action policy and value through the action network and value network, as depicted in Fig.~\ref{pipeline}.

\subsubsection{Observation Space $O$}


\begin{figure}[!t]
\centering
\includegraphics[width=0.45\textwidth]{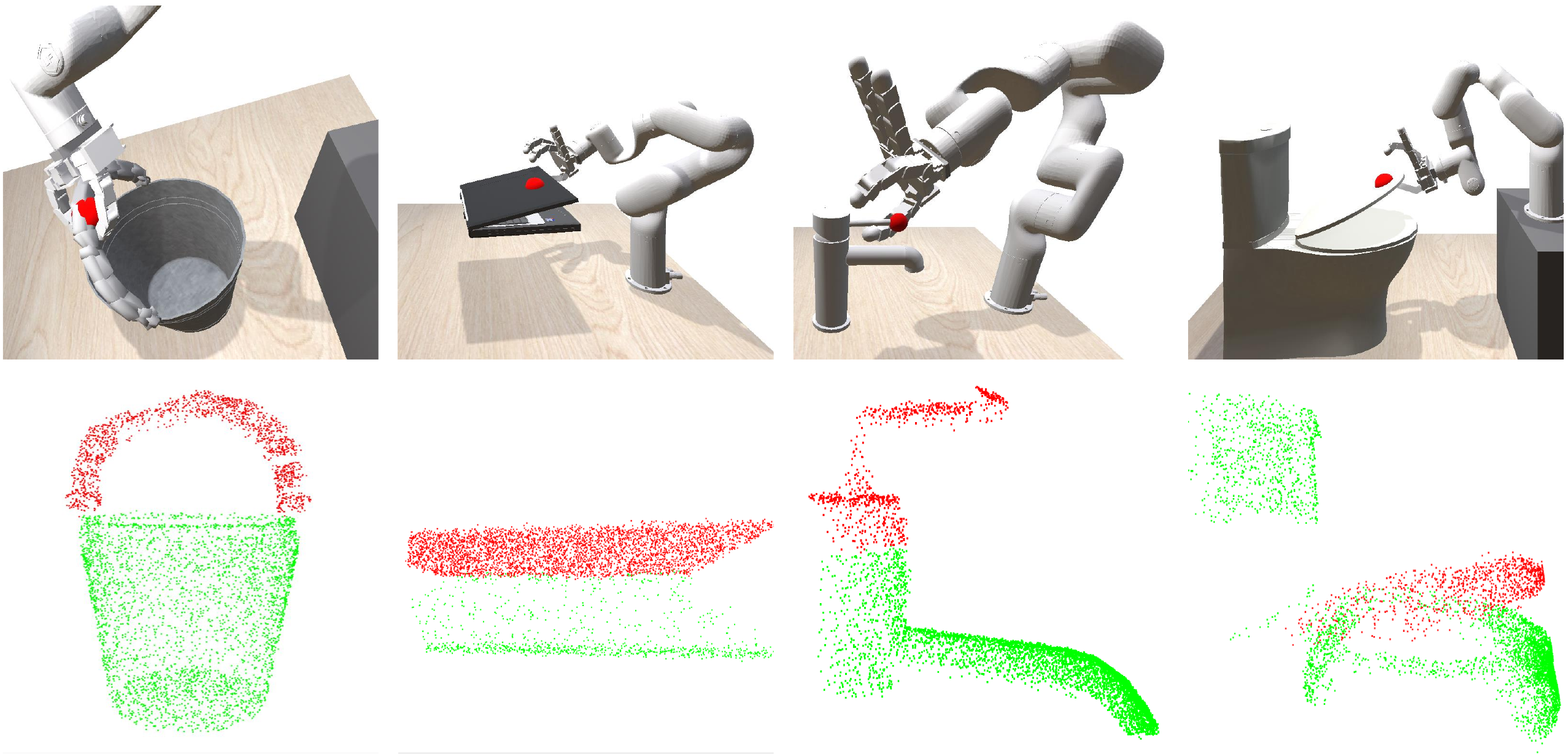}\\
\caption{The center of the red ball indicates the position the finger should approach, as shown at the top of the figure. It will adjust as the functional part of the object changes. Below, the illustrations depict the segmentation results for the functional parts across four tasks.}
\label{target_and_segmentation_point_cloud}
\end{figure}


Our observation space is composed of three modalities:
\begin{itemize}
    \item \textbf{Point Cloud $\mathcal{P}_2$.} Captured from the reinforcement learning (RL) camera, as depicted in Fig.~\ref{pipeline}, this point cloud is used to extract features from the entire scene. The feature extraction is performed using a simplified version of PointNet.
    \item \textbf{Robot Proprioception $\mathcal{S}_r$.} This modality provides proprioceptive data, aiding the model in predicting the next action based on the current state.
    \item \textbf{Object Goal Position and Rotation.} This modality includes the goal position and rotation of the object for each trial. Fig.~\ref{target_and_segmentation_point_cloud} illustrates the goal position for each task.
\end{itemize}

\textbf{Feature Extractor Pre-training. }To enhance the model's understanding of the scene, we construct a pre-training dataset for the point cloud, distinct from the segmentation dataset. This dataset comprises four categories: the functional part of the object, the non-functional part of the object, the robotic arm, and the dexterous hand. For each category, we generate 6,000 point clouds. We then train PointNet on this dataset, and the resulting pre-trained network is used to initialize the feature extractor for PPO.

\subsubsection{Action Space $A$}

Our action space is represented by a 22-dimensional vector, encompassing both the 6 DoF of the robotic arm and the 16 DoF of the dexterous hand, corresponding to the dimensions of the dexterous grasp pose $\mathcal{G}$. The 6-dimensional vector captures the angular and linear velocities of the robotic arm's end-effector in three spatial directions. Using this 6-dimensional vector and the scene's update time step, we compute the translation and rotation of the robotic arm's end-effector. Subsequently, we employ IKFast to calculate the joint angles of the robotic arm. The 16-dimensional vector represents the joint angles of the Allegro hand, which are controlled via Proportional-Derivative (PD) control. Finally, we scale and clip the 22-dimensional vector to ensure it remains within the allowable operational range.

\subsubsection{Reward Function $R^\prime$}
Given the initial dexterous grasp pose, the objective is to guide the dexterous hand in interacting with the object and driving the tasks to completion. To achieve this, it is essential to ensure that the robot completes the task both safely and efficiently. We decompose the reward into three components: interaction reward $r_{in}$, completion reward $r_{co}$ and restriction reward $r_{re}$. The overall reward function is defined as follows:

\begin{equation}
    R^\prime = \alpha r_{in} + \beta r_{co} + \eta r_{re}.
\end{equation}

Here, the $\alpha, \beta, \eta$ are balance parameters. The reward function is designed to be dense, ensuring task completion within a limited timeframe, and general enough to be applicable across various tasks.

\textbf{Interaction Reward. }
To successfully complete the task, it is crucial to generate effective interactions between the dexterous hand and the object. Emulating human manipulation requires a reward structure that encourages finger dexterity, with the palm serving an auxiliary role. We define the interaction reward $r_{in}$ as follows:

\begin{equation}
    r_{in} = r_{finger} + r_{palm},
\end{equation}
where
\begin{align}
\begin{split}
    r_{finger} = \sum_{X_i \in X} min(-\Vert X_i - X_{object} \Vert, \lambda),
\end{split}\\
\begin{split}
    r_{palm} = min(\Vert Y_{palm} - Y_{object} \Vert, \gamma),
\end{split}
\end{align}

Here, the $r_{finger}$ encourages the dexterous hand to manipulate the object using its fingers, while $r_{palm}$ ensures that the proper amount of force is applied to the object. The $X$ denotes the position of each finger of the dexterous hand, while $X_{object}$ refers to the position of the center of the red ball in the object's functional part (as shown in Fig.~\ref{target_and_segmentation_point_cloud}). The $Y_{palm}$ and the $Y_{object}$ are the rotation of the palm and  the object's functional part, respectively. The terms $\lambda$ and $\gamma$ are the regularization terms  designed to prevent abrupt changes in the reward function.

\textbf{Completion Reward. }
The primary objective is to ensure task completion during interaction. Therefore, a reward is provided when the dexterous hand successfully manipulates the object by adjusting the initial dexterous grasp pose. To encourage quicker task completion, the reward is designed as:
\begin{equation}
    r_{co} = Progress(task) + \delta \  IsCompletion  (H_s - H_c),
\end{equation}
where $Progress(task)$ is the evaluation function that measures the current task progress, and $IsCompletion \in \{0, 1\}$ indicates whether the task has been completed. $H_s$ represents the maximum exploration step, and $H_c$ denotes the current exploration step during the exploration process. The parameter $\delta$ serves as a balancing factor. For instance, the progress of opening a laptop can be quantified by the opening angle.

\textbf{Restriction Reward. }
During object manipulation, the robot may perform unstable or unsafe actions, such as excessive joint speeds or collisions between the robotic arm and the object. To mitigate these undesired behaviors, we introduce a restriction reward, which is divided into two components: the reward to limit the robot's own movements $r_{robot}$ and the reward to manage interactions between the robot and the environment $r_{en}$. The restriction reward $r_{re}$ is defined as:
\begin{equation}
    r_{re} = r_{robot} + r_{en},
\end{equation}
where
\begin{align}
\begin{split}
    r_{robot} = -(\Vert G_{t+1, palm} - G_{t, palm} \Vert + q_{vel}),
\end{split}\\
\begin{split}
    r_{en} = -\sum_{j_i \in J} IsCollision(j_i, object).
\end{split}
\end{align}

Here, $G_{t+1, palm}$ and the $G_{t, palm}$ denote the 6-dimensional pose of the palm at time steps $t+1$ and the $t$, respectively, while $q_{vel}$ refers to the joint angle velocity of the robotic arm. The set $J$ represents the joint mesh of the robotic arm, the $IsCollision \in \{0, 1\}$ is used to detect collisions between the robotic arm and the object.

\section{EXPERIMENTS}
We begin by conducting extensive simulation experiments to evaluate our dexterous manipulation method across four tasks: opening a laptop, opening a toilet, turning on a faucet and lifting a bucket. Our method demonstrates significant improvements in success rates across all tasks compared to the baseline. To further validate our approach, we deploy our method in a real-world environment, where it continues to show strong performance, particularly in opening a laptop and lifting a bucket.

\subsection{Simulation Experiments}
\subsubsection{Simulation Experimental Setup}

We trained our model using the SAPIEN physical simulator with the XArm6 robot arm and Allegro Hand, which have 6 DoF and 16 DoF, respectively. The scene time step and control frequency were set to 0.004s and 50 Hz, respectively, with a friction coefficient of 5. The test settings mirrored those of the training phase. The objects used in the simulation are sourced from the PartNet-Mobility dataset ~\cite{xiang2020sapien} and are divided into train and test datasets in a 3:2 ratio. These objects are scaled, transformed, and randomly placed within the robotic arm’s workspace, with an additional random rotation applied along the plane of the table. The point cloud data captured by the RL camera is downsampled to 512 points, with Gaussian noise added to enhance the realism of the simulation.

\subsubsection{Baseline Algorithm and Ablation Study}

Our method for manipulating objects with dexterous hands is based on an initialized dexterous grasp pose. To establish a baseline, we compared our approach against three algorithms:
\begin{itemize}
    \item \textbf{Manipulation from a Fixed Position:}  This approach assumes no prior knowledge of the object’s position and serves as an equivalent to the DexArt method.
    \item \textbf{Manipulation from a Given Position $\mathbf{t}$ (MGP):}  This algorithm incorporates prior knowledge of the initial position only, demonstrating the significance of selecting an appropriate initial position.
    \item \textbf{Manipulation from a Given Position $\mathbf{t}$ and Initial Rotation $\mathbf{R}$ (MGPR):}  This method incorporates both the initial position and the initial manipulation view, highlighting the importance of these factors in achieving effective manipulation.
\end{itemize}

\begin{figure}[htbp]
    \centering
    \subfigure[bucket]{\includegraphics[width=0.22\textwidth]{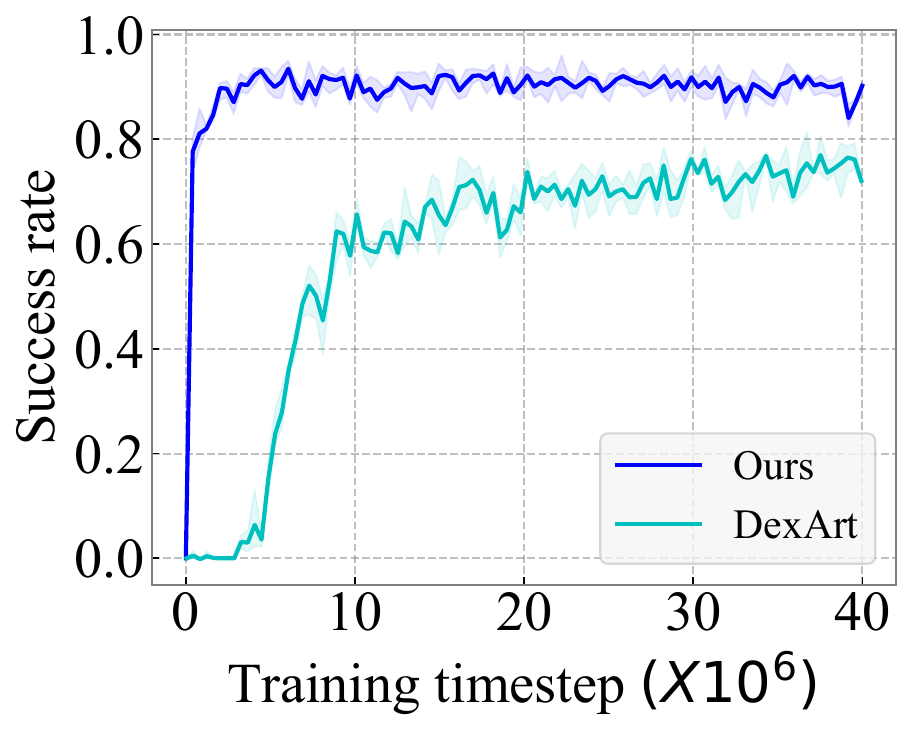}}
    \subfigure[laptop]{\includegraphics[width=0.22\textwidth]{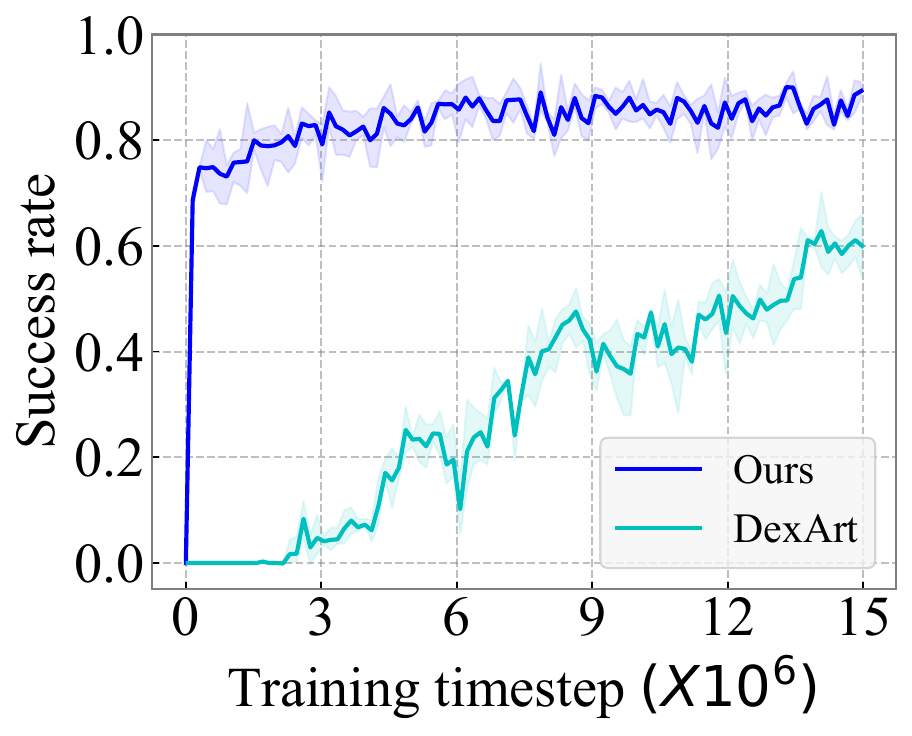}}
    \subfigure[faucet]{\includegraphics[width=0.22\textwidth]{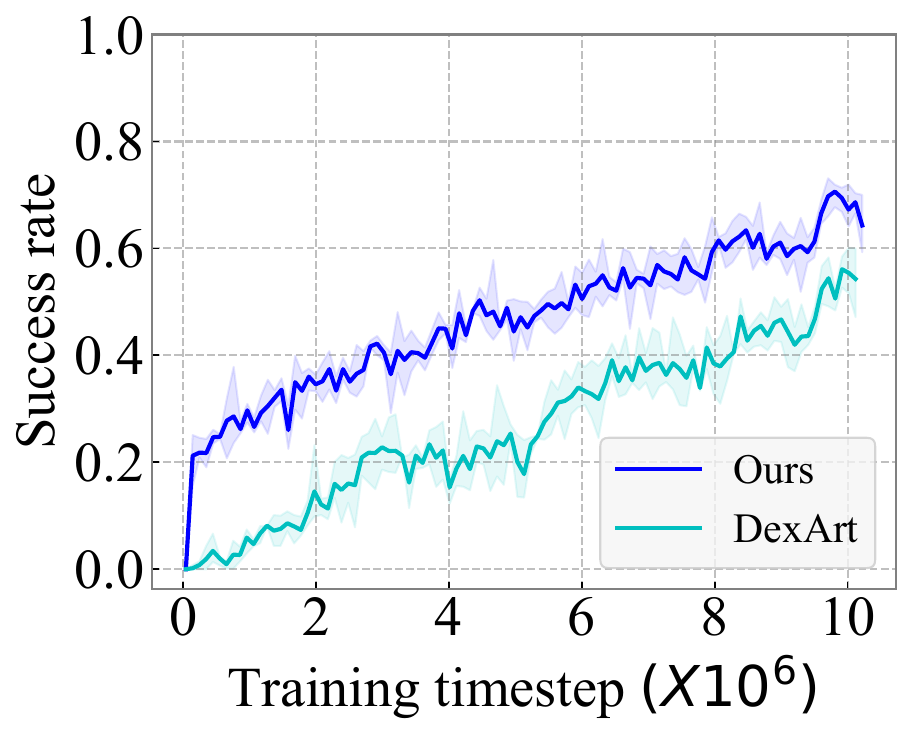}}
    \subfigure[toilet]{\includegraphics[width=0.22\textwidth]{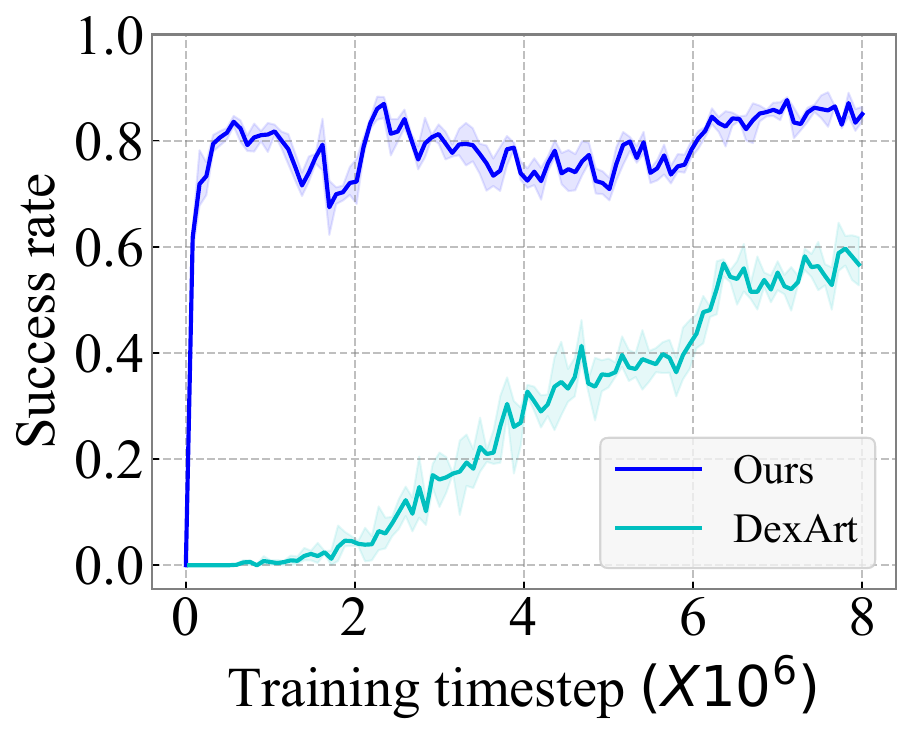}}
    \caption{Illustration of the success rate as a function of the training process using XArm6 on the simulation test dataset. }
    \label{main_result}
\end{figure}

\begin{table}[!t]
\centering
\begin{tabular}{c|cccc}
\hline
\textbf{Category} & Bucket & Laptop & Faucet & Toilet\\
\hline
\textbf{Ours} & \textbf{94.17} & \textbf{90} & \textbf{73.33} & \textbf{87.88}\\
\hline
\textbf{Ours(big)} & 85.21 & 76.3 & 47.73 & 42.38\\
\hline
\textbf{DexArt} & 79.17 & 66.67 & 60 & 58.57\\
\hline
\textbf{DexArt(big)} & 54.58 & 34 & 17.56 & 34.29\\
\hline
\textbf{MGPR} & 89.02 & 87.33 & $\times$ & $\times$\\
\hline
\textbf{MGP} & 87.26 & 79.33 & $\times$ & $\times$\\
\hline

\end{tabular}
\caption{The success rate (\%) of four tasks on the simulation test dataset. 'big' represents that we expend the test space by a factor of 16.}
\label{simulation result}
\end{table}

\subsubsection{Main Results}

We evaluate our model using three random seeds and report the average success rate, with each object being tested 100 times. The test workspace is identical to the training workspace. All results are summarized in Table.~\ref{simulation result}, with the training process illustrated in Fig.~\ref{main_result}. Our method achieves a 75\% success rate at a time step of $0.4\times10^6$ on the bucket-lifting task, 65\% at $0.1\times10^6$ on the laptop-opening task, and 62\% at $0.1\times10^6$ on the toilet-opening task. In contrast, the success rate of the DexArt method is nearly 0\% at $1.5\times10^6$. Additionally, our method reaches a 92\% success rate at $5\times10^6$ time steps on the bucket-lifting task, 87\% at $6\times10^6$ on the laptop-opening task, and 87\% at $2.5\times10^6$ on the toilet-opening task. In summary, our method enhances the success rate by 15\% to 29.31\% and improves learning efficiency by a factor of 80 to 150 compared to DexArt. Although the improvement in success rate for the faucet-turning task is not as pronounced as in the other three tasks, our method still demonstrates a significant increase in success rate during both the initial and final stages of training.

\begin{figure}[htbp]
    \centering
    \subfigure[bucket]{\includegraphics[width=0.22\textwidth]{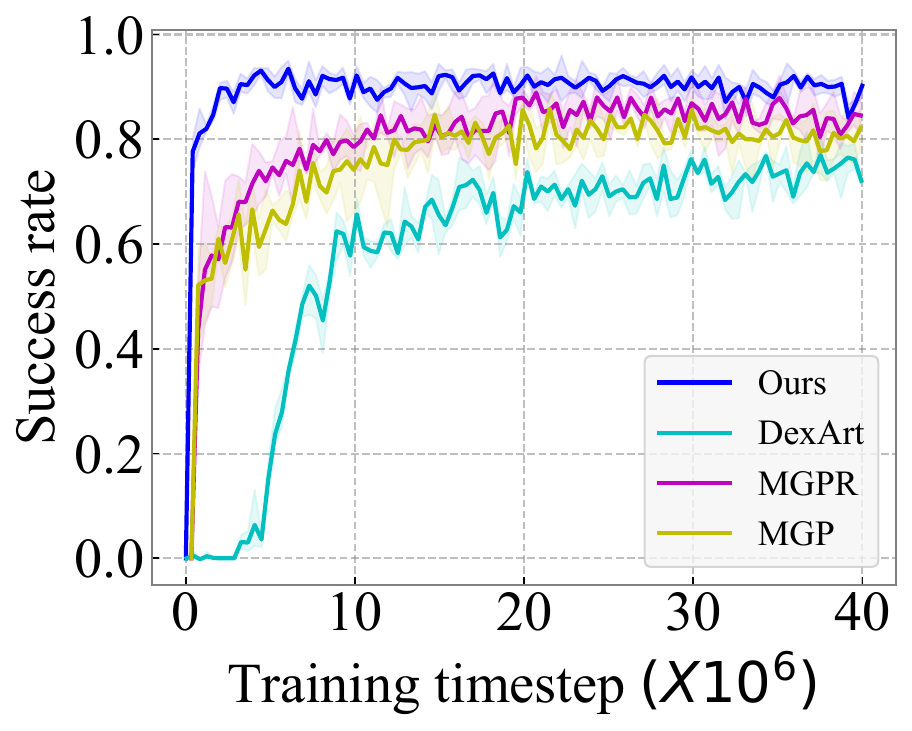}}
    \subfigure[laptop]{\includegraphics[width=0.22\textwidth]{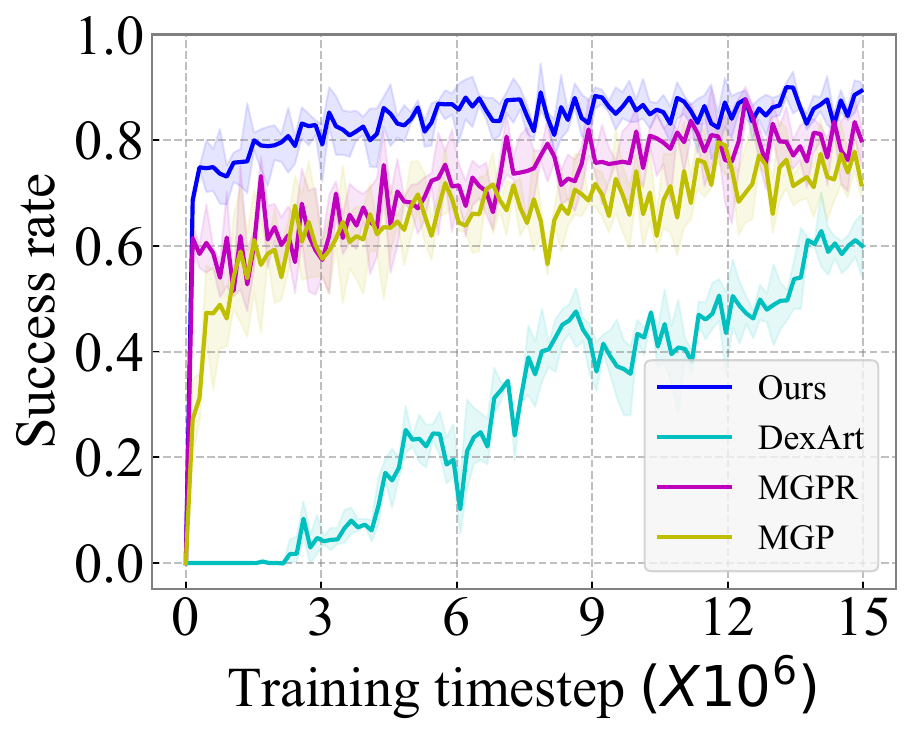}}
    \caption{ Comparison of success rates between our method and various exploration policies using XArm6 on the simulation test dataset.}
    \label{ablation_result}
\end{figure}

To further investigate the impact of different components of the dexterous grasp pose on reinforcement learning, we evaluate the MGPR and MGP methods on the bucket-lifting and laptop-opening tasks. Fig.~\ref{ablation_result} illustrates the success rates as a function of training time steps. The results indicate that both the position $\mathbf{t}$ and rotation $\mathbf{R}$ contribute to improved performance across all tasks. When incorporating only the prior knowledge of position $\mathbf{t}$, the success rate reaches 87\% at $20\times10^6$ time steps for the bucket-lifting task and 79\% at $12\times10^6$ time steps for the laptop-opening task. By adding prior knowledge of rotation $\mathbf{R}$, the success rate increases to 89\% at $20\times10^6$ for the bucket-lifting task and 87\% at $13\times10^6$ for the laptop-opening task. Including the prior knowledge of grasp $\mathbf{B}$ further enhances the success rates to 94\% and 90\%, respectively. These findings demonstrate that as prior knowledge increases, learning efficiency improves significantly. Without this prior knowledge, success rates are notably lower. Additionally, the manipulation process with prior dexterous grasp poses knowledge more closely resembles human actions, such as using fingers to grasp and lift the bucket handle, rather than relying on the palm as in DexArt. For detailed comparison videos, please refer to the supplementary materials.

\begin{figure}[htbp]
    \centering
    \subfigure[faucet]{\includegraphics[width=0.22\textwidth]{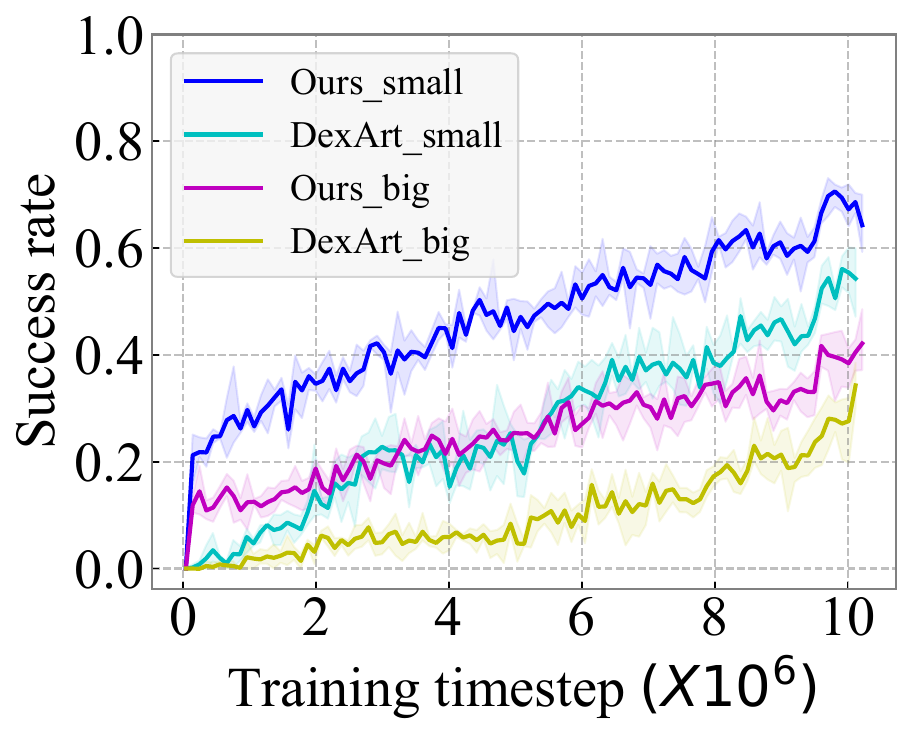}}
    \subfigure[toilet]{\includegraphics[width=0.22\textwidth]{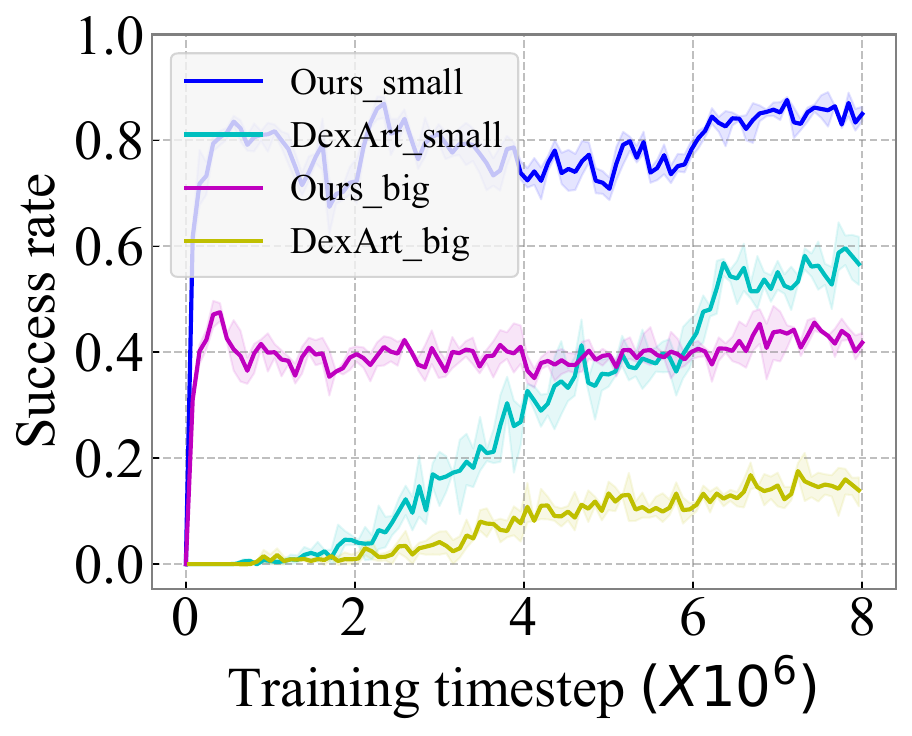}}
    \caption{Comparison of success rates between our method and DexArt using XArm6 on the simulation test dataset.}
    \label{scope_result}
\end{figure}

To further evaluate our model, we expand the test space by a factor of 16 to challenge the model with extreme object placements. Objects are randomly placed within this enlarged workspace to assess whether the model could successfully complete the tasks. As with previous evaluations, the model is tested three times, and the average success rate is calculated. While the success rates decline in this more challenging environment, our model still demonstrates a significant advantage over the baseline, confirming the effectiveness of incorporating prior dexterous grasp pose knowledge. Table.~\ref{simulation result} and Fig.~\ref{scope_result} present the final results and the training process, respectively.

\begin{figure*}[!t]
\centering
\includegraphics[width=0.98\textwidth]{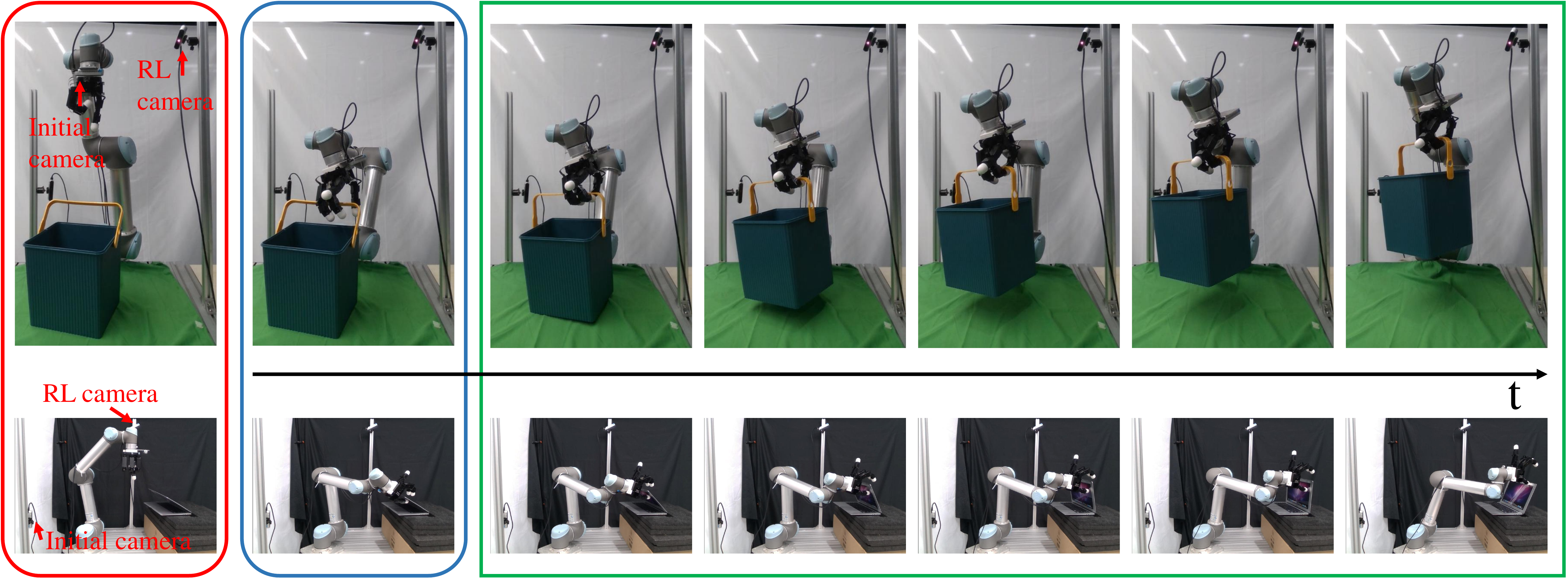}\\
\caption{The red box illustrates the real-world setting, the blue box shows the initial dexterous grasp pose, and the green box shows the manipulation process over time in the real world.}
\label{ur5_manipulation_process}
\end{figure*}
\subsection{Real-world Experiments}
\begin{figure}[htbp]
    \centering
    \subfigure[bucket]{\includegraphics[width=0.22\textwidth]{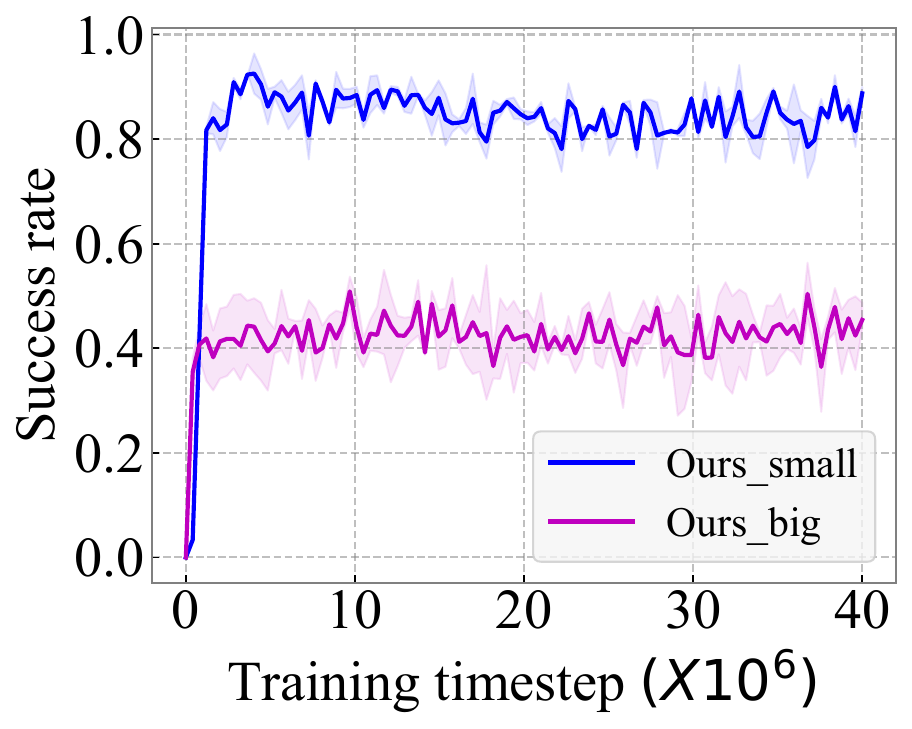}}
    \subfigure[laptop]{\includegraphics[width=0.22\textwidth]{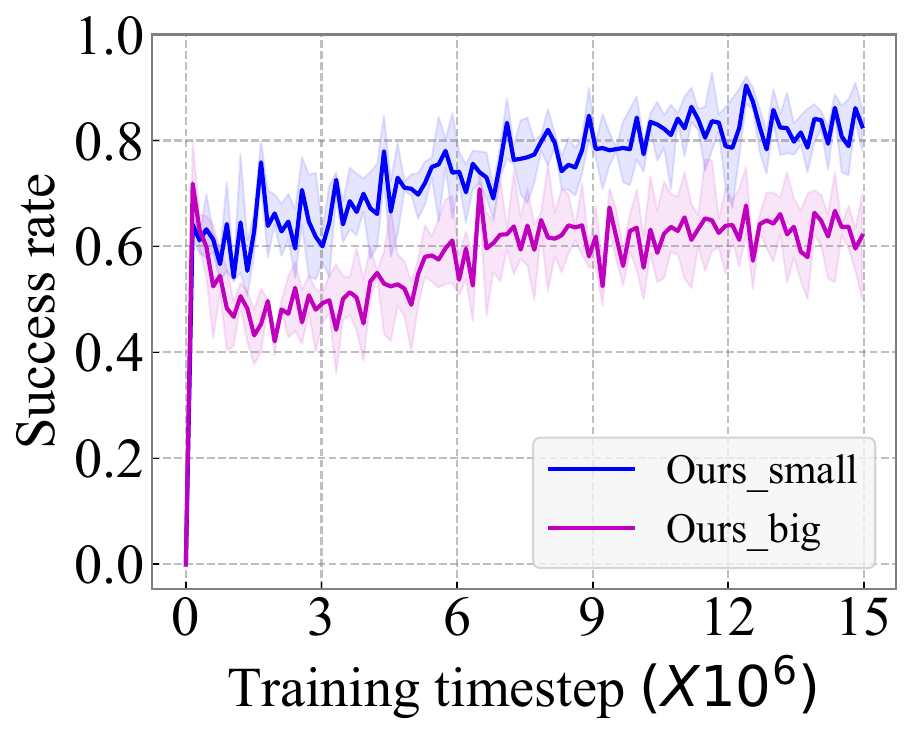}}
    \caption{Comparison of success rates between our method and DexArt using the UR5 on the simulation test dataset.}
    \label{ur5_training_process}
\end{figure}
To validate our model in a real-world environment, we select two tasks: opening a laptop and lifting a bucket. The experimental setup is illustrated in Fig.~\ref{ur5_manipulation_process}. The Allegro Hand is mounted on the end of a UR5 robotic arm, with a D415 camera providing visual input. For the laptop-opening task, we place foam boards under the workspace to elevate it. This setup not only prevents collisions between the dexterous hand and the tabletop but also ensures that the robotic arm does not move into unreachable positions. For the bucket-lifting task, a nail is inserted into the side of the bucket to prevent the handle from falling over during the manipulation.

\begin{figure}[!t]
\centering
\includegraphics[width=0.45\textwidth]{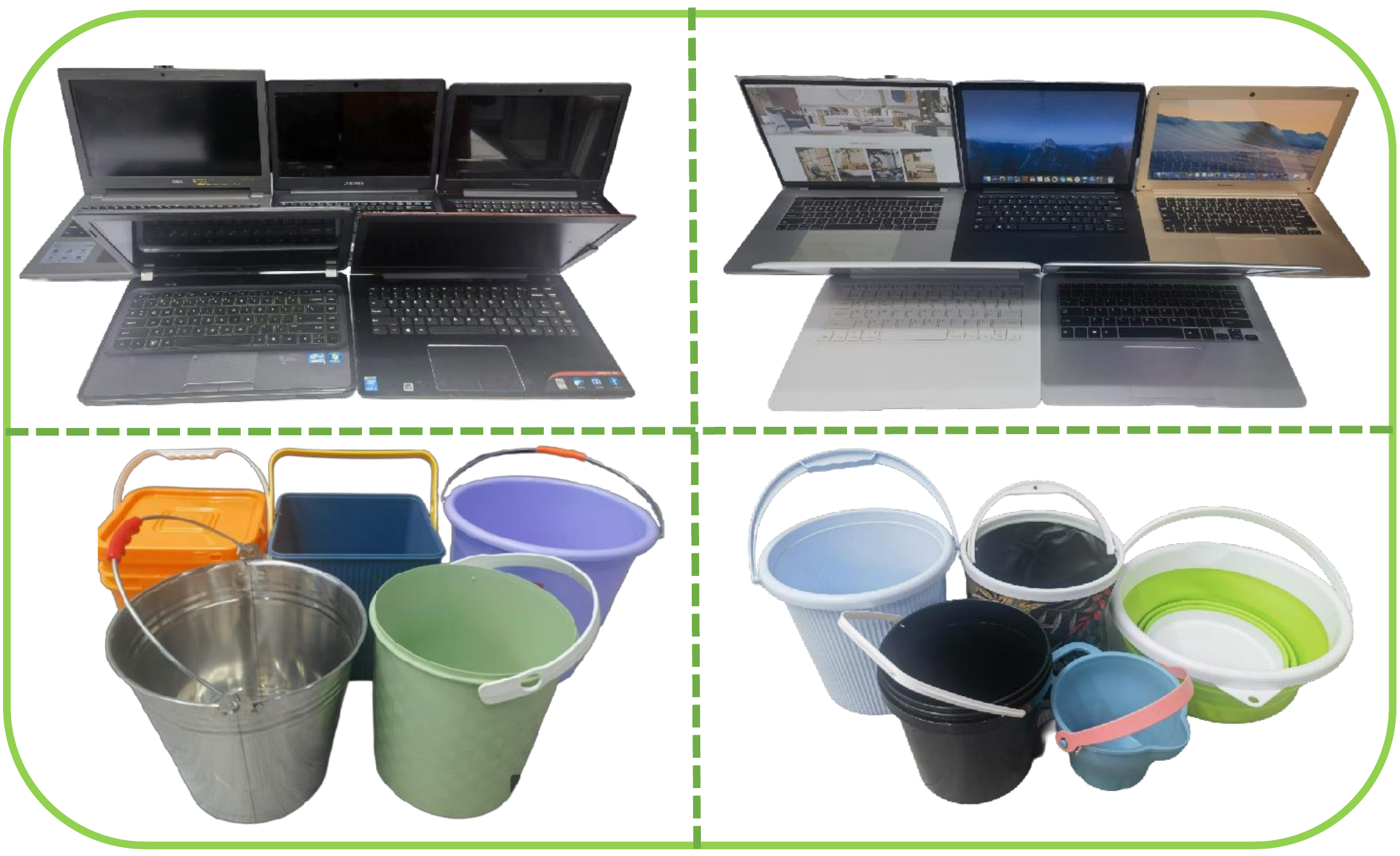}\\
\caption{Test objects in the real world. On the left are the more difficult objects for the laptop and bucket tasks, while on the right are the easier objects.}
\vspace{-0.15cm}
\label{ur5_test_object}
\end{figure}


Fig.~\ref{ur5_test_object} displays all the test objects used in the real-world experiments. The laptops are categorized into two types: one is a model designed for decorative purposes, which is easier to open, while the other is a real, scrapped laptop that is more difficult to operate due to its stiffness. The bucket test set includes items made from various materials, sizes, and qualities commonly found in daily life. We classify the lighter and smaller buckets as simpler objects, while the heavier and larger buckets are considered more challenging to lift.


To adapt our model for real-world experiments, we first reconstruct the segmentation dataset and point cloud pre-training dataset using the UR5 robotic arm. These datasets are used to train the segmentation model and feature extractor. We then retrain our PPO model in a simulation environment with camera poses identical to those in the real-world setup. Fig.~\ref{ur5_training_process} illustrates the training process for these two tasks, and the results are consistent with those observed using the XArm6. In other words, our method is independent of robotic arm systems and exhibits a certain degree of generalizability. For the bucket-lifting task, a successful outcome is defined as lifting the bucket to a height of at least 12 cm. Since the robotic arm is fixed to the table in the real world, excessive lifting height could lead to collisions between the robotic arm and the bucket.

\begin{table}[!t]
\centering

\label{ur5_test_result}
\begin{tabular}{cccc}
\hline
\textbf{Category} & Easy & Difficult & Average\\
\hline
\textbf{Bucket(real)} & 84.62 & 73.58 & 79.08\\
\hline
\textbf{Laptop(real)} & 77.19 & 71.15 & 74.31\\
\hline
\textbf{Bucket(sim)} & $\times$ & $\times$ & 93.75\\
\hline
\textbf{Laptop(sim)} & $\times$ & $\times$ & 90\\
\hline
\end{tabular}
\caption{'real' represents the success rate (\%) of opening laptop and lifting bucket in the real world. 'sim' represents the success rate (\%) of these tasks on simulation test dataset.}
\label{real_world_result}
\end{table}

Fig.~\ref{ur5_manipulation_process} illustrates the manipulation process of the dexterous hand in a real-world environment. An object is randomly placed in the workspace, and the process is tested approximately 10 times. The results of these two tasks are summarized in Table.~\ref{real_world_result}. The findings suggest that failures in the laptop-opening task are often caused by the incorrect force exerted by the robotic arm, leading to the palm becoming stuck against the laptop lid or the robotic arm getting stuck against the foam board. In the bucket-lifting task, failures are typically due to the hand not grasping the handle properly or the bucket being too heavy to lift.


\section{CONCLUSION}
This paper presents a novel approach to dexterous manipulation by leveraging prior knowledge of dexterous grasp poses. Our method is rigorously evaluated through a series of simulations and real-world experiments across various tasks, including lifting a bucket, opening a laptop, turning on a faucet, and opening the toilet. The results consistently demonstrate that incorporating prior grasp pose knowledge significantly enhances both learning efficiency and success rates, outperforming baseline methods in both simulated and real-world settings.

The success of our approach underscores the importance of combining reinforcement learning with structured prior knowledge, particularly in complex manipulation tasks that require fine motor control. Despite the challenges encountered, especially in real-world applications, the findings indicate that our method holds substantial promise for practical deployment in robotic systems.

Future work will focus on addressing the limitations observed in real-world experiments, such as refining force control and improving the adaptability of the model to various object types and environments. Additionally, we aim to explore further enhancements in the integration of visual and tactile feedback to enable more human-like dexterity in robotic manipulation.
\printbibliography
\end{document}